\documentclass[11pt]{article}

\usepackage{graphicx,graphics}
\usepackage[authoryear]{natbib}
\usepackage{graphics,amsmath,color}
\usepackage{helvet,setspace}

\usepackage[margin=1in]{geometry}

\begin{document}

\title{The ``given data'' paradigm undermines both cultures}

\author{Tyler H. McCormick\footnote{tylermc@uw.edu, thmccormick.github.io}\\
       University of Washington}

\date{}

\maketitle
\vspace{-10pt}
\begin{abstract}
Breiman organizes \emph{Statistical modeling: The two cultures} around a simple visual.  Data, to the far right, are compelled into a ``black box'' with an arrow and then catapulted left by a second arrow, having been transformed into an output.  Breiman then posits two interpretations of this visual as encapsulating a distinction between two cultures in statistics.  The divide, he argues is about what happens in the ``black box.''  In this comment, I argue for a broader perspective on statistics and, in doing so, elevate questions from ``before'' and ``after'' the box as fruitful areas for statistical innovation and practice.
\end{abstract}


\paragraph{Note:} This comment was prepared as part of a special issue to appear in \emph{Observational Studies} on the 20th anniversary of Leo Breiman's article entitled \emph{Statistical modeling: The two cultures}.~\\~\\

 \doublespacing

I sincerely appreciate the opportunity to comment on the twentieth anniversary of Breiman's \emph{Statistical modeling: The two cultures}.  Along with the ``two cultures,'' Breiman also identifies another divide in his Rejoinder, this time between an era of great potential for statistics and a worry by many statisticians about the future of the discipline.  I don't think this dissonance has gone away in the twenty years since Breiman's article and, if anything, there's an argument that the rise of the (generally) more ecumenical moniker of ``data science'' and the proliferation of easy to use tools for complex analysis should compel even more foreboding in response to Breiman's article.  

I, however, am optimistic.  I think the comparative advantage of statistics is more salient than ever.  To see this positive perspective, though, we need a broader view of our discipline than is apparent from Breiman's framing.  Breiman describes the distinction between the ``two cultures'' in statistics as being essentially about what happens in the ``black box'' between input data and outputs.  Under Breiman, so-called algorithmic modelers fill the black box with methods focused on prediction, whereas data modelers purport to summarize nature.    
%
%
%
Focusing only on the box relegates statistics to a ``given data'' discipline where there's a (false) partition between active exploration of methodology and its application to passive, fixed data.  
If ``given data'' statistics were grammar, ``data'' would be a direct object, subject to the whims of the transitive verb ``statistical methodology.''
In the data modeling culture, data arise from a particular set of distributional assumptions.  The data themselves may be infinite draws but the \emph{distributions} are fixed by the modeler.  The algorithmic culture makes no such assumptions but routinely benchmarks an algorithm by comparing its performance with others on the same dataset. 
In either case, there's a sense that once data have been collected, whether the design and data quality are perfect or highly flawed, the primary opportunity to enhance scientific understanding is by further interrogating said data with (typically) increasingly complicated methods. In contrast, I see many of the most exciting opportunities in our discipline as happening ``before'' or ``after'' the box and, in terms of comparative advantage, I see statisticians as uniquely poised to be leaders in these settings.  

I'll give a couple of examples from my own work to make this more concrete.  First, in my work in global health I work frequently on estimation problems in settings where data are incredibly sparse (e.g. estimating cause of death distributions in settings where most deaths happen outside of hospitals and aren't formally recorded).  The ``given data'' paradigm leaves us with two unsatisfactory options: (i) use whatever data sources are readily available and develop increasingly complex models, either data or algorithmic, to extrapolate beyond the extremely limited available data or (ii) wait until full coverage, high quality data are available.  While the first option is sometimes valuable, it also creates opaque risks as extrapolation carries unknown and, potentially arbitrarily large, errors.  Particularly given the complexity of modern models, we can't expect estimates to be valuable by simply ``handing them off'' to policymakers.  ~\citet{boerma2018monitoring} echos this idea, pointing out, ``{global public health experts and academics are often oblivious to the limitations of these estimates and regularly overinterpret the numbers, especially if the estimates support their arguments}.''  

Why are statisticians uniquely poised to make an impact in this setting? I contend that it is because the most pressing questions in this setting require understanding how to thoughtfully use limited resources for new data collection (``before'' the box) and how to faithfully communicate to policymakers the quality of information in the estimates (``after'' the box).  Areas of statistics like survey sampling and data visualization no doubt play critical roles here.  The deeper connection, though, goes back to fundamental practical and philosophical questions at the core of statistical thought.  Given imperfect and highly limited data, should the value of new data collection be tied inherently to making decisions (e.g. allocating interventions or resources)?  Are subjective expert beliefs admissible as evidence?  What qualifies as sufficient evidence for a change in the burden of a particular disease? How do we combine evidence from two studies that use different designs? These questions echo core philosophical debates in the field and remain whether we use an algorithmic or modeling perspective, with of course important caveats about the implications thereof.       
 
In my first example, I argue for more emphasis on ``before'' and ``after'' the black box.  What if, though, the process didn't proceed neatly from inputs to analysis to outputs?  The twenty years since Breiman's piece have seen explosive growth in measurement technology across multiple disciplines, along with creative and resourceful use of passively observed data sources (e.g. administrative records from courts or electronic health records).  Though there's great potential for these data sources, they're often not fundamentally designed for research, or at least not our specific research questions.  One basic implication of this opportunism is that we now need to invest substantial time and energy in transforming and representing the data to fit into our algorithmic or data models.  Applied regression modelers know this process well, no doubt.  A recent trend has also introduced, however the prospect of \emph{predicting} data that go into algorithmic or data model.  Taking my work in global health again as an example, if a person dies outside of the hospital, a survey with a family member might provide inputs for a classifier that predicts a likely cause of death.  A subsequent analysis takes these \emph{predicted} causes of death and then fits (for example) regression models to understand how cause of death varies by age, geography, access to care, or other salient factors.  Over the course of this very common exercise we've used both the algorithmic and data modeling frameworks, one to predict the outcome and another to make descriptive statements about variation in the object we've predicted.  Regardless of how we think about the prediction step, the second phase of modeling is fundamentally about quantifying and propagating uncertainty.  Rather than only uncertainty about the sampling distribution and inference model, though, we now also have uncertainty that comes from the prediction process (both in terms of how the observations in the training data were collected and in the specification of the predictive model).  The context may be a bit different from many standard inference problems, but the objective couldn't be more statistical.    


To conclude, I'd like to shift the conversation, in statistical practice, research, and education, to focus more on ``before'' and ''after'' the box.  My ``outside the box'' (apologies for the image that's now in your mind of a florescent light flickering off a trite motivational poster just beyond the adjacent cubicle) is certainly not unique, as any of the themes I've touched on here exist both in literature in statistics and in other fields.  It is also to expand, rather than undermine, the ``two cultures'' paradigm, as the themes presented here are consequential whether we use algorithmic modeling, data modeling, or (as is increasingly the case) both.  Fundamentally, my conclusion is the same as Breiman's.  Namely that if ``we define the boundaries of our field in terms of familiar tools and familiar problems, we will fail to grasp the new opportunities.''

\paragraph{Acknowledgements.} Research reported in this publication was supported by the National Institute Of Mental Health of the National Institutes of Health under Award Number DP2MH122405. The content is solely the responsibility of the authors and does not necessarily represent the official views of the National Institutes of Health. 








\bibliographystyle{abbrevnamed}
\bibliography{extrapolate.bib}

\end{document}